\documentclass[letterpaper, 10 pt, conference]{ieeeconf}
\IEEEoverridecommandlockouts
\usepackage{capt-of,etoolbox}
\usepackage{amsmath,amssymb,amsfonts}
\usepackage{graphicx}
\usepackage{textcomp}
\usepackage{multirow}
\usepackage{xcolor}
\usepackage{algorithm}
\usepackage{algpseudocode}
\usepackage{amsmath}
\usepackage{hyperref}
\usepackage{afterpage}  
\usepackage{placeins}
\usepackage{tabulary}
\usepackage[utf8]{inputenc}
\usepackage{booktabs} 
\usepackage{wrapfig}
\usepackage[
    style=ieee,
    doi=false,
    isbn=false,
    url=false,
    eprint=false,
    backend=bibtex,
    natbib=true
    ]{biblatex}
\usepackage{float}
\newcommand{\acro}{\textsc{OUTDOOR}}

\title{\LARGE \bf Reasoning about the Unseen for Efficient Outdoor Object Navigation\\
}

\author{
    Quanting Xie$^{1}$\hspace{2em} Tianyi Zhang$^{1}$\hspace{2em} Kedi Xu$^{1}$\hspace{2em} Matthew Johnson-Roberson$^{1}$\hspace{2em} Yonatan Bisk$^{1}$}

\begin{document}



\twocolumn[{%
\renewcommand\twocolumn[1][]{#1}%
\maketitle
\begin{center}
    \centering
    \includegraphics[width=0.93\textwidth]{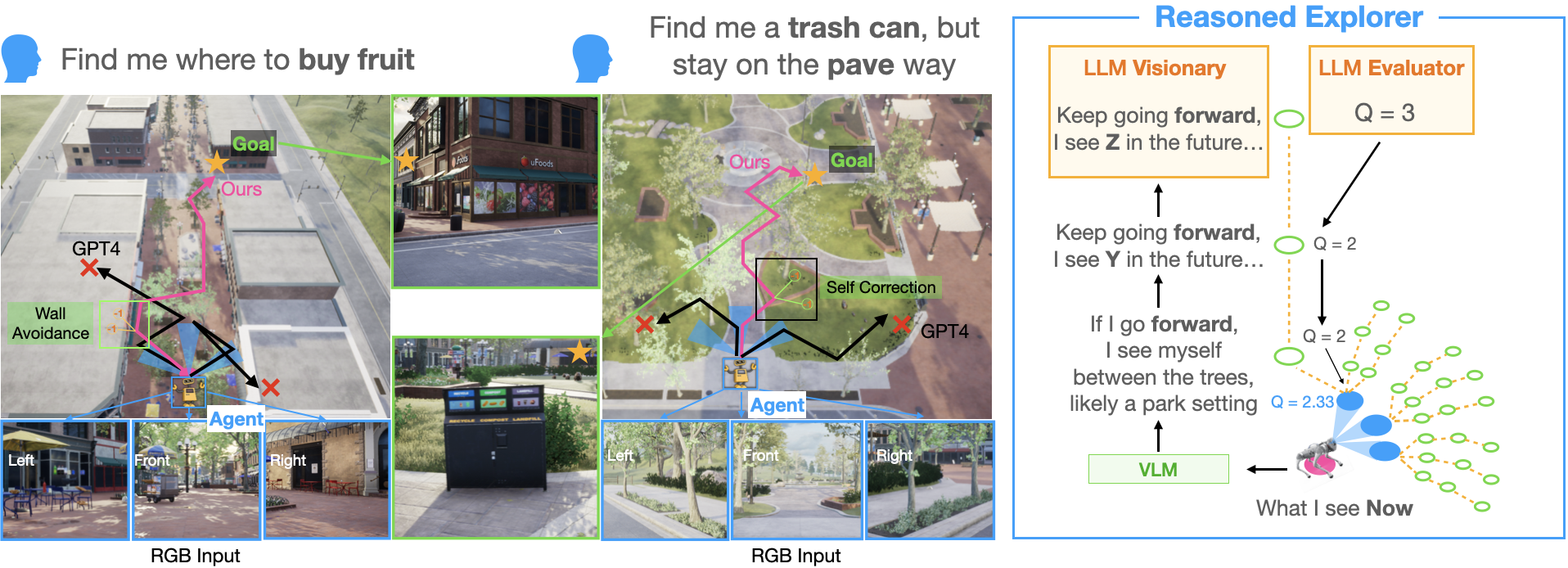}
    \captionof{figure}{Direct application of Language Models in embodied agents navigating outdoor environments suffers from short-sightedness and limited environment comprehension. Our approach augments the LLM by enabling it to expand imaginary nodes in space, enhancing feasibility for outdoor navigation.}
    \label{fig:overview}
\end{center}%
}]

\begin{abstract}
Robots should exist anywhere humans do: indoors, outdoors, and even unmapped environments.
In contrast, the focus of recent advancements in Object Goal Navigation (OGN)\cite{anderson2018objectnav, chaplot2020object,majumdar2022zson} has targeted navigating in indoor environments by leveraging spatial and semantic cues that do not generalize outdoors. While these contributions provide valuable insights into indoor scenarios, the broader spectrum of real-world robotic applications often extends to outdoor settings. As we transition to the vast and complex terrains of outdoor environments, new challenges emerge. Unlike the structured layouts found indoors, outdoor environments lack clear spatial delineations and are riddled with inherent semantic ambiguities. Despite this, humans navigate with ease because we can reason about the unseen.  We introduce a new task \textbf{\acro}, a new mechanism for Large Language Models (LLMs) to accurately hallucinate possible futures, and a new computationally aware success metric for pushing research forward in this more complex domain. Additionally, we show impressive results on both a simulated drone and physical quadruped in outdoor environments. Our agent has no premapping and our formalism outperforms naive LLM-based approaches. 
\end{abstract}
\let\thefootnote\relax\footnote{$^{1}$Carnegie Mellon University, Pittsburgh, PA USA. {\tt\small  \{quantinx, tianyiz4, kedix, mkj, ybisk\}@andrew.cmu.edu}}
\let\thefootnote\relax\footnote{$^{2}$\href{https://github.com/quantingxie/ReasonedExplorer}{https://github.com/quantingxie/ReasonedExplorer}}

\section{Introduction}
\label{sec:intro}

Advancements in Object Goal Navigation (OGN)~\cite{anderson2018objectnav, chaplot2020object,majumdar2022zson} have enhanced the proficiency of robotic agents in navigating indoor environments by leveraging spatial and semantic cues. Agents that can guide humans (e.g. the visually impaired~\cite{blind-indoors}) are an important enabling technology, but need to move beyond restricted indoor spaces to the full richness of outdoor navigation.
Outdoor environments are substantially larger than handled by current semantic mapping approaches~\cite{chaplot2020learning,Min2022}, have complex terrains~\cite{self-supervised-outdoor}, and, crucially, lack clear semantic delineations. 
Not only is sensing simplified indoors, but so is reasoning as rooms are easily distinguished and semantically categorized.  Outdoor environments still have semantic distinctions but visually identical spaces might be a soccer field, a picnic area, or the pit of an outdoor orchestra depending on the time of day. Additionally, outdoor navigational tasks typically demand that robotic agents engage in roles with more granular goal specifications. For instance, in the context of search and rescue operations, the objective is not merely to navigate to a general category of `people' but to pinpoint casualties potentially trapped under a car. 

Recently, Large Language Models (LLMs)~\cite{gpt, devlin2018bert, openai2023gpt4} trained on expansive internet datasets are serving as adaptable policies in embodied platforms, making them proficient in addressing a wider range of tasks~\cite{saycan2022arxiv, shah2023lmnav, brohan2022rt}. The existing work has primarily focused on high-level task-planning with predefined skills in constrained environments. 
Despite this, we have seem very promising skill demonstrations in indoor object-scenarios made possible by these models~\cite{chen2023train, zhou2023esc, zhou2023navgpt}.
While some emergent behavior has been identified, the language and vision communities have begun harnessing these LLMs for their reasoning capabilities due to the vast world knowledge and models stored in their parameters. 

Thus, we posit that outdoor navigation offers a promising avenue to test and refine the foundational navigation and reasoning abilities of LLMs. This paper aims to formulate an elementary navigation policy and evaluate its efficacy in diverse and challenging outdoor environments, providing insights into the potential of LLMs as embodied agents.

Our primary contribution in this work are: 
\begin{enumerate}
    \item We introduce the \textbf{\acro{}} {\small (\textbf{O}utdoor \textbf{U}nderspecified \textbf{T}ask \textbf{D}escriptions \textbf{O}f \textbf{O}bjects and \textbf{R}egions)} task, which dramatically increases the complexity inherent in object goal navigation for outdoor settings.
    \item We introduce a novel use of LLMs as a planning agent to traverse real-world outdoor terrains. Our approach imagines future notes for a RRT (Rapidly-exploring Random Tree) to improve agent success (+50.4\%).
    \item We introduce the \textbf{CASR} {\small (\textbf{C}omputationally \textbf{A}djusted \textbf{S}uccess \textbf{R}ate)} metric, that trades off planning costs with time spent ``thinking" (i.e. querying LLMs).
\end{enumerate}

\section{Task Definition}
\label{sec:taskdefination}
\begin{figure}[t]
    \centering
    \includegraphics[width=1\linewidth]{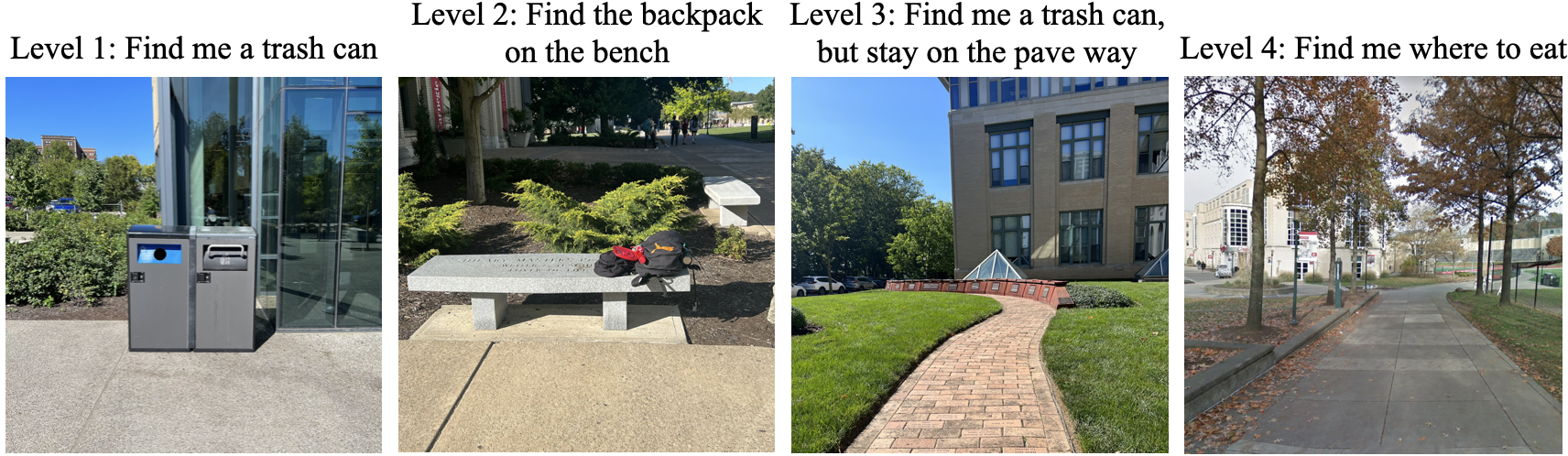}
    \caption {Above are example queries at varying levels of complexity and a representative scene in our \textbf{\acro} task.}
    \label{fig:explore}
\end{figure}

\subsection{\textsc{OUTDOOR}: Outdoor Underspecified Task Descriptions Of Objects and Regions }

In traditional Object Goal Navigation, 
users specify distinct goal categories that can be automatically evaluated by the system and do not include contraints or handle underspecified goals (e.g. reference by affordance). However, real-world outdoor environments like parks present more complex scenarios. For example, if the goal is to find a place to eat, it could refer to any bench or table, rather than a specific one. \textbf{\acro} embraces ambiguity, generalizing to a more nuanced and realistic navigational challenge.
We categorize the instruction complexity into four levels:

\begin{enumerate}
    \item \textbf{Level 1}: Navigate to obj X \textcolor{gray}{[aka traditional object-nav]}
    
    \item \textbf{Level 2}: Navigate to obj X conditioned on obj Y
    
    \item \textbf{Level 3}: Navigate to obj X conditioned on path P

    \item \textbf{Level 4}: Navigate to underspecific abstraction A
\end{enumerate}

Human intervention is essential for evaluating success across all levels from 1 to 4. While the goals in Levels 1 to 3 are specific and relatively straightforward to assess, Level 4 presents a more abstract goal. For example, a directive like "Find me somewhere to take a nap?" makes the evaluation more nuanced, potentially necessitating 
human 
evaluation.
Agents start an episode from a pose \( s_0 \) and are given a linguistic goal \( x \) from one of the aforementioned levels. The agent's challenge is to reconcile real-time environmental observations with its interpretation of \( x \) and understand the semantic and spatial relationships between the objects and regions present. Operating autonomously, the agent must then navigate the environment, with the path being represented as \( \{s_0, a_0, s_1, a_1, \ldots, s_T, a_T\} \), where each action \( a_t \) transitions to a pose \( s_{t+1} \).
The episode terminates when the agent predicts ``Found Goal". Agents are also limited to a maximum exploration time: \(T_{max}\).


\section{Related Works}
\label{sec:relatedwork}

\subsection{Decision Making and Planning for LLM}
Vanilla implementations of Large Language Models (LLMs) often fall short in decision-making and planning capabilities. To address this, several strategies have been developed. The linear reasoning approach, ``Chain of Thoughts," enhances structured problem-solving~\cite{wei2023chainofthought}, and tree-based strategies, ``Tree of Thoughts," bring forth search-guided reasoning capabilities~\cite{yao2023tree}. To improve performance search algorithms have also been integrated~\cite{xie2023decomposition}.

External planning methods have emerged~\cite{hao2023reasoning,zhao2023large,zhang2023planning, wang2023planandsolve, wang2023describe, huang2022inner, yao2023react} as methods to leverage techniques such as Monte Carlo Tree Search (MCTS) to enhance the reasoning capacities of LLMs. While they show promise in fields like mathematics, code generation, and high-level task planning, their application in low-level path planning for robotics remains limited. A primary reason is the complexity of mapping LLM outputs to the intricate action spaces of robots, making tasks requiring detailed sequences of movements a challenge. In this context, our approach stands out by using waypoints as a natural interface for low-level path planning. This not only bridges the gap between LLM reasoning and robot actions but also ensures that our method operates in a parallel manner, offering both effectiveness and efficiency.

\subsection{LLM as embodied agents for navigation}
The use of language for guiding embodied agents has a long lineage.  Language only models like 
BERT~\cite{devlin2018bert} can be used as scoring functions between language instructions and path to help embodied agent navigate~\cite{majumdar2020vlnbert}. Performance on such tasks have scaled with larger models (e.g. GPT-4) which have greater aptitudes for common sense reasoning and its comprehension of world structures 
~\cite{zhou2023esc, chen2023train, zhou2023navgpt, shah2023lmnav, dorbala2023catshapedmug}. 
Shah et al ~\cite{shah2023lmnav} uses LLMs as a parser to extract landmarks as sub-goal nodes for robots to navigate on a graph, Chen et al~\cite{chen2023train} use LLM as an evaluator to re-weight the waypoints generated by the frontier-based method ~\cite{1997frontier}. Zhou et al ~\cite{zhou2023esc} took another approach, they used several hand-designed constraints via Probabilistic Soft Logic programming language to choose the best frontiers to explore. NavGPT~\cite{zhou2023navgpt} utilized the synergizing prompt methods such as ReAct~\cite{yao2022react} with a discrete action space for LLM to navigate. 

\begin{figure*}[!t]
    \centering
    \includegraphics[width=0.9\textwidth]{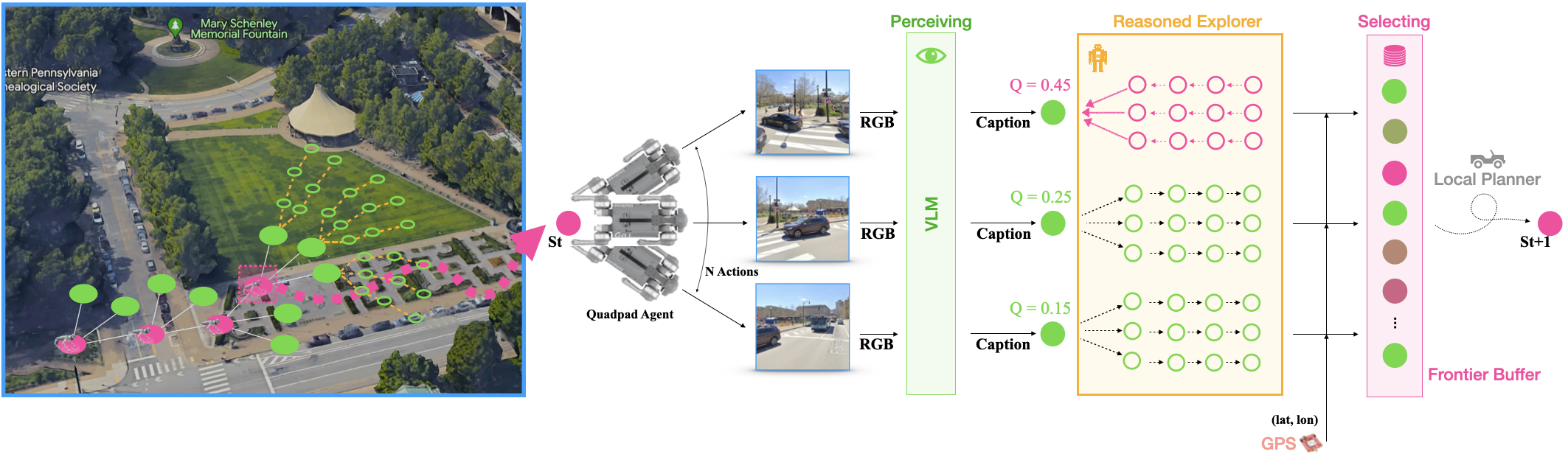}
    \captionof{figure}{Overview: The agent captures \(N\) RGB images (potential frontiers). Each image is processed through a Vision Language Model (VLM) to generate a textual caption. Subsequent Rapidly-exploring Random Trees (RRT) aid the agent in envisioning possible future scenarios for each frontier. The results, combined with GPS coordinates, populate a frontier buffer. The most promising frontier is identified, and a local planner guides the agent to its location.}
    \label{fig:overview}
\end{figure*}

Existing approaches to navigation predominantly rely on idealized indoor scene graphs, often assuming structured environments where, for example, a refrigerator is necessarily located in the kitchen or a fireplace in the living room ~\cite{zhou2023esc, chen2023train, shah2023lmnav}. Alternatively, some methods leverage Google Street View data for navigation tasks ~\cite{schumann2023velma}. However, these approaches fall short in capturing the nuanced complexities and granularities inherent to real-world scenarios, such as search and rescue operations or advanced domestic robotics tasks in semantically rich environments such as airports or campus buildings. In genuine outdoor settings, spatial semantics may be ambiguous or lack well-defined boundaries. Consequently, an intelligent agent must be capable of strategically  predict information in space to effectively navigate and reason within these more complex contexts.

\section{Method: Reasoned Explorer}
\label{sec:method}

Figure 3 outlines our proposed \textit{Reasoned Explorer} method -- an LLM reasoning technique that enables an LLM-based agent to execute \textbf{\acro} tasks in complex outdoor environments. We remove the perfect depth assumption and use a dynamically expandable graph to store the map information illustrated in \S \ref{sec:graph_frontier}. We then employ two LLMs (\S \ref{sec:llm_mcts}): one as a visionary and the other as an evaluator. The visionary LLM is designed to project future agent states and potential scenarios, while the evaluator critically assesses the feasibility of achieving the goal within those states. To physically embody our method, we then talks about the perception and action techniques used in \S \ref{sec:vlm} -- \ref{sec:action}

\subsection{Graph the unknown}
\label{sec:graph_frontier}

Historically, methods for object goal navigation and VLN relied on near perfect depth information derived from simulations to generate dense geometric occupancy maps, which subsequently informed the expansion of explorable frontiers~\cite{chen2023train,zhou2023esc, zhou2023navgpt,pmlr-v155-anderson21a}. However, these assumptions falter in real-world outdoor settings, especially without the aid of high-end depth cameras or LiDARs. Addressing this limitation, our approach introduces an adaptive topological graph to introduce frontiers. This combination not only mitigates the need for perfect depth information but also enhances the agent's navigational capabilities.

As depicted in Figure \ref{fig:explore}, the green circles symbolize the expanded frontiers emanating from pathpoints, which are denoted by pink circles. During each iteration, the algorithm calculates and expands \(N\) frontiers from the present pathpoint. These frontiers subsequently undergo a rigorous planning and scoring phase, as elaborated in Section \ref{sec:llm_mcts}. All of these frontiers are retained in a specialized \texttt{Frontier Buffer} for subsequent reference.
To ensure that distant frontiers are penalized, yet remain viable for exploration, we introduce a sigmoid-modulated distance function:
\[
\sigma(d_i) = \frac{1}{1 + e^{-k(d_i - d_0)}}
\]
Here, \( \sigma(d_i) \) represents the sigmoid-modulated distance for the \(i\)-th frontier. The parameter \( k \) dictates the sharpness of the modulation, with a larger \( k \) creating a more pronounced transition around \( d_0 \), which represents the distance where the penalty is half its maximum potential value.
The agent's actions are then determined by the updated score function:
\[
S_{t+1}^* = \arg\max_{i} \left( Q(S_{t+1}) - \sigma(d_i) \right)
\]

Within this formulation,\(Q(S_{t+1})\) refers to the score of each frontier following the planning-scoring phase. As the agent advances in its exploration, the selected frontier — now deemed a pathpoint \(S_{t+1}^*\) — is removed from the \texttt{Frontier Buffer}.
The agent persists in its exploration endeavors until it perceives a halt signal dispatched from the a speicalized LLM checking function.

\begin{figure}[h]
    \centering
    \includegraphics[width=0.92\linewidth]{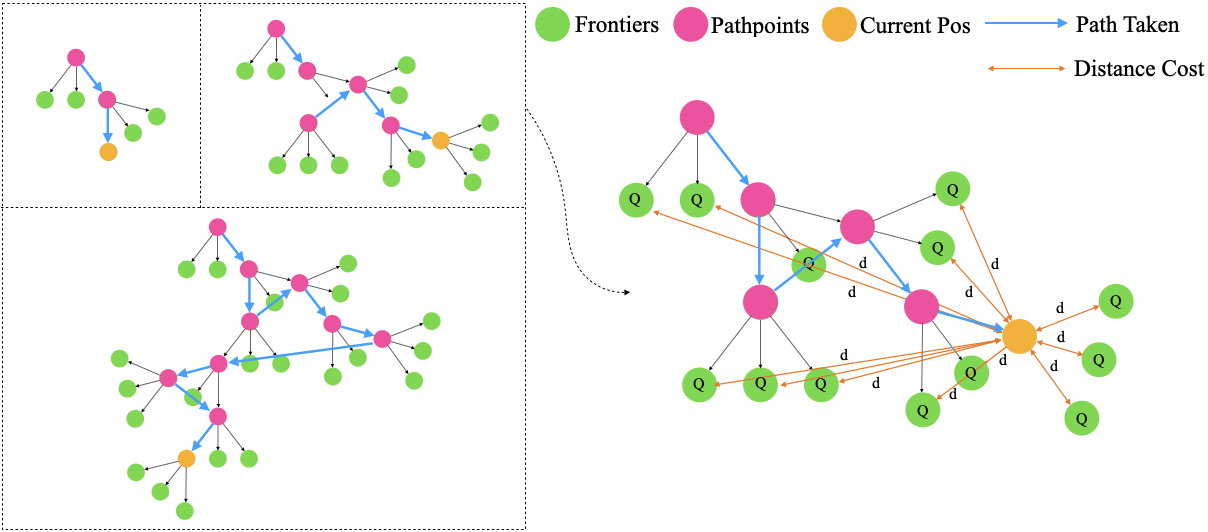}
    \caption{The left image illustrates the expansion process where, at each step, \(N\) nodes are expanded (with \(N = 3\) as depicted). The right image shows the agent's decision-making process with distance cost at each step. }
    \label{fig:explore}
\end{figure} 

\subsection{Reasoning about the uncertainty}
\label{sec:llm_mcts}
In the context of planning within intricate outdoor environments, direct determinations by the LLM-based method solely on localized information can lead to suboptimal behaviors, such as inconsistency and short-sightedness. These behaviors arise not only from the inherent tendency of LLMs to hallucinate~\cite{McKenna2023SourcesOH}, but also the non-delineated nature of the outdoor environment. Such properties become particularly concerning when placing heavy reliance on singular output generations from LLMs~\cite{stochastic}.

To fortify against these vulnerabilities and enhance decision robustness, we elected to incorporate the future information using expanding RRT strategy. By projecting multiple forward-looking imaginary branches through iterative queries of LLM\_Visionary and LLM\_Evaluator (Figure \ref{fig:overview}), mitigating the risks associated with single query outputs. Our choice of RRT was informed by its intrinsic properties, in contrast of the sequential sampling process used in MCTS\cite{zhang2023planning}\cite{hao2023reasoning}, RRT provides a parallelizable framework to allow us score and expand the imaginary nodes all at the same time. 

\begin{figure}[t]
    \centering
    \includegraphics[width=1\linewidth]{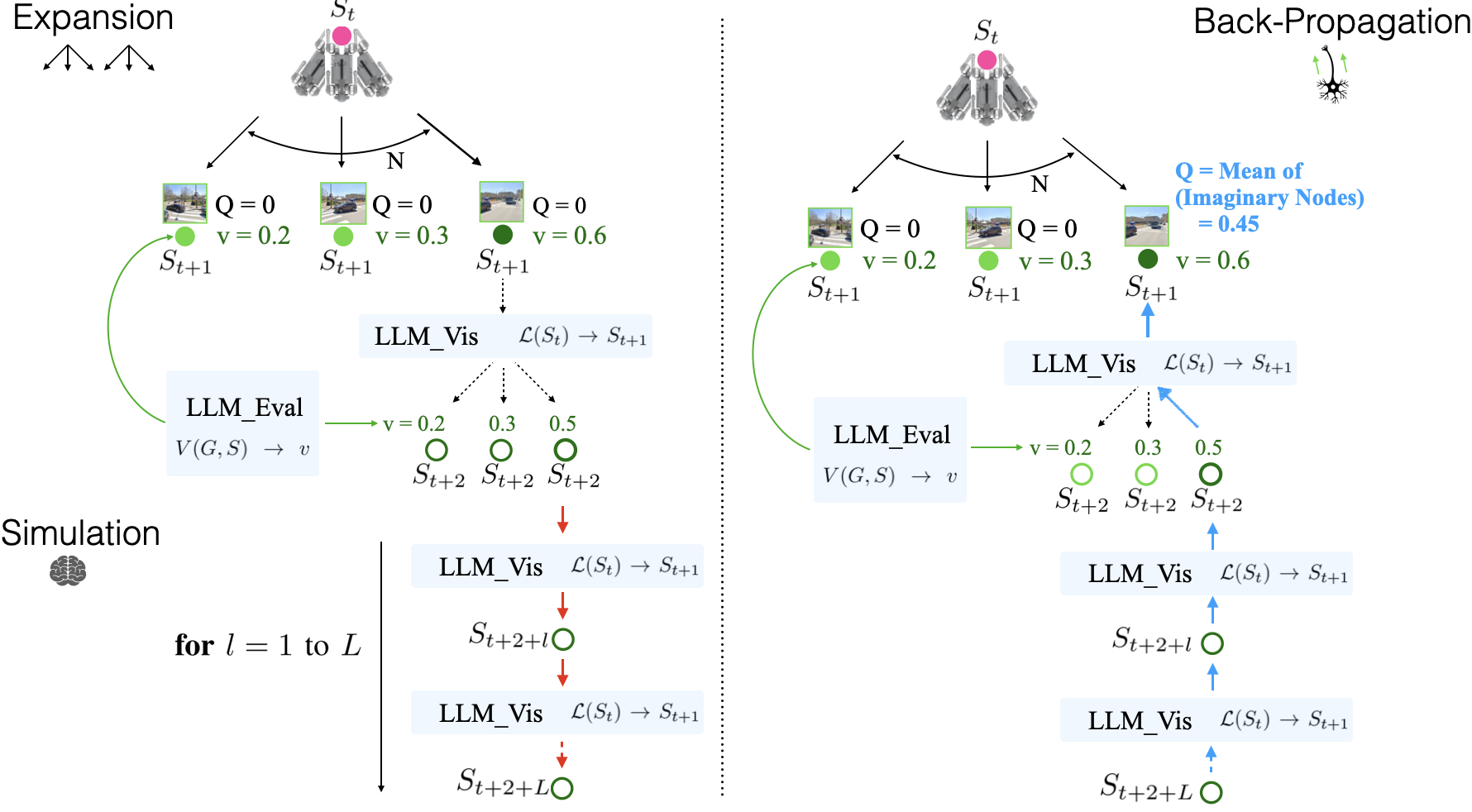}
    \caption{The left image illustrates the expansion process where, at each step, \(N\) nodes are expanded (with \(N = 3\) as depicted). The right image shows the agent's decision-making process with distance cost at each step. }
    \label{fig:rrt}
\end{figure}

In our integration of the RRT, the specifics and underlying mechanics are meticulously outlined in Algorithm 1 and visually complemented in Figure \ref{fig:rrt}. At its core, our adaptation hinges on the dual roles assumed by the LLM: as an agent, denoted as \( \text{LLM\_Eval} \), and as an evaluator, represented by \( \text{LLM\_Gen} \). Their specific responsibilities and interplay will be further discussed in the following sections.

\paragraph{LLM\_Evaluator} 
denoted as \( V(G, S) \rightarrow v \), assesses the correlation between a scene description \( S \) and the provided goal objects or instructions \( G \). Its primary role is to guide the agent by offering a reference score, indicating the likelihood of achieving the goal based on the current scene. The scoring mechanism is structured on a Likert scale ranging from 1 to 5, where a score of 1 indicates a low likelihood of goal achievement, and a score of 5 signifies a high likelihood. This evaluative approach ensures that the agent can make informed decisions based on the contextual relevance of the scene to the goal.

\paragraph{LLM\_Visionary} 
represented as \( \mathcal{L}(S_t) \rightarrow S_{t+1} \), produces the next scene descriptor \( S_{t+1} \) based on the current scene description \( S_t \). Its objective is to enable the agent to anticipate or predict its future waypoint. This is achieved by prompting the agent to envision what it might encounter next. Comprehensive prompt templates and further details are available on our official website.

In our adaptation of the RRT, the algorithm's mechanics are determined two hyperparameters:

\begin{itemize}
    \item \textbf{N}: Dictates the action space dimension, signifying the range of feasible directions the agent can embark upon during its exploration.
    
    \item \textbf{L}: Denotes the length of individual simulations within the branch. This dictates the depth of the tree and how far the algorithm projects into potential future states.
\end{itemize}

\begin{algorithm} [t]
\caption{LLM-RRT}
\begin{algorithmic}[1]
\vspace{5pt}
\Procedure{LLM-RRT}{$K, S_{tlist}$}
    \State $S_{t+2}^{sample} \leftarrow \Call{Expansion}{S_{t+1}^*}$
    \State $v_{mean} \leftarrow \Call{Simulation}{S_{t+2}^{sample}}$
    \State $Q(S_{t+1}^*)\leftarrow $\Call{Back-propagation}{$v_{mean}, depth$}
\EndProcedure
\end{algorithmic}
\end{algorithm}

\subsection{Perceiving}
\label{sec:vlm}

The agent captures \(N\) images during each exploration steps. These images are processed by a Vision Language Model (VLM). For the purposes of this study, we employed Kosmos-2 \cite{kosmos-2}, a VLM fine-tuned with spatially-grounded image-text data. This model offers the distinct advantage of providing detailed object-level descriptions of scenes. Importantly, it is promptable, so we prompt it to describe not only the objects in the scene, but also the spatial relationship between objects as well as the backgrounds.

\subsection{Action on real robot}
\label{sec:action}
Upon waypoint determination by the scoring function detailed in Section \ref{sec:graph_frontier}, our robot employs a straightforward PID controller to traverse the path navigating between the current waypoint and the subsequent one, with the localization from on board high resolution RTK-GPS and IMU.

\section{Experiments}
\label{sec:exp}
\begin{table*}[h]
\centering
\begin{tabular}{@{}lc@{\hspace{5pt}}c@{\hspace{5pt}}c@{\hspace{5pt}}c@{\hspace{5pt}}c@{\hspace{20pt}}
                    c@{\hspace{5pt}}c@{\hspace{5pt}}c@{\hspace{5pt}}c@{\hspace{5pt}}c@{\hspace{20pt}}
                    c@{\hspace{5pt}}c@{\hspace{5pt}}c@{\hspace{5pt}}c@{\hspace{5pt}}c@{\hspace{20pt}}
                    c@{\hspace{5pt}}c@{\hspace{5pt}}c@{\hspace{5pt}}c@{\hspace{5pt}}c@{}}
                    \toprule
& \multicolumn{5}{c}{SR} 
& \multicolumn{5}{c}{OSR} 
& \multicolumn{5}{c}{SPL} 
& \multicolumn{5}{c}{CASR} \\ 
            & L1   & L2   & L3 & L4 & Avg        & L1   & L2   & L3 & L4 & Avg & L1   & L2   & L3 & L4 & Avg  & L1   & L2   & L3 & L4 & Avg \\ 
\cmidrule{2-5}
\cmidrule{7-10}
\cmidrule{12-15}
\cmidrule{17-20}
    LLM-as-Eval & 0.17 & 0.09 & 0.00 & 0.00 & 0.06     & 0.17 & 0.27 & 0.00 & 0.00 & 0.06 & 0.13 & 0.04 & 0.00 & 0.00 & 0.04 & 0.10 & 0.06 & 0.00 & 0.00 & 0.04 \\
LLM-MCTS(10 iter)    & 0.54   & 0.43   & 0.59   & \textbf{0.33}   & 0.47         & \textbf{0.88}   & 0.76   & 0.63   & 0.69   & 0.74   & 0.37   & 0.31   & 0.38   & \textbf{0.23}   & 0.32   & 0.11   & 0.09   & 0.11   & 0.07   & 0.10   \\
\midrule
Ours & \textbf{0.59} & \textbf{0.49} & \textbf{0.63} & 0.32 & \textbf{0.51} & \textbf{0.88} & \textbf{0.82} & \textbf{0.71} & \textbf{0.88} & \textbf{0.82} & \textbf{0.44} & \textbf{0.32} & \textbf{0.43} & 0.22 & \textbf{0.35} & \textbf{0.51} & \textbf{0.42} & \textbf{0.46} & \textbf{0.28} & \textbf{0.42} \\
\bottomrule
\end{tabular}
\caption{Baseline Comparison for Different Task Levels in simulation (AirSim~\cite{shah2017airsim}}
\vspace{-10pt}
\label{tab:success}
\end{table*}

Comprehensive evaluations are conducted across multiple platforms: 1. The AirSim~\cite{shah2017airsim} simulation environment: A photo-realistic outdoor simulation setting that consists of different semantic distinguishable areas in Downtown West environment. 2. A real-world robotic platform: Unitree Go1, equiped with a USB camera, high resolution RTK-GPS module, and Inertial measurement unit. 

\subsection{A Compute Aware Metric for LLM-based Robotic Agents}
\label{sec:optimization}

In the assessment of robotic agents in real-world settings, particularly for outdoor tasks like Search and Rescue, it is vital to strike a balance between navigation efficiency, computational overhead, and travel duration. The dominant metrics in the space:  Success Rate (SR) and Success Weighted by Path Length (SPL), ignore ``time".  Here, we specifically mean wallclock time, or the length of an experiment or episode.  While always relevant in practical scenarios, the use of Large Foundation Models (e.g. over API) introduces a new computational trade-off.  Specifically, the interplay between Computational Time (CT) and Travel Time (TT). An increase in computation that results in reduced travel time might lead to overall efficiency gains depending on the amount of computational time required.  Colloquially, \textit{when is it faster to think before acting, versus acting on a hunch?}

\paragraph{Formulating the CASR Metric}
This overall efficiency versus the maximum allowed episode length is simply a normalized sum of the two components: Compute (CT) and Travel (TT) time.  The value is normalized to the range [0,1], aligning it for integration with the traditional SR metric:

\begin{equation}
I_{CT, TT} = 1 - \frac{CT + TT}{T_{max}}
\end{equation}

\( T_{max} \) is a predefined maximum acceptable time for an mission to be completed, in our experiments, \( T_{max} \) = 30 minutes, and any experiment time above it is set to failure. 

With the normalized SR and the aforementioned interaction term, we formulate the Computationally Adjusted Success Rate (CASR) as:

\begin{equation}
CASR = SR \times \left( \frac{1}{N} \sum_{i=1}^{N} I_{CT_i, TT_i} \right)
\end{equation}

The range of CASR spans [0,1], where:

\begin{itemize}
    \item 1 signifies optimal performance, reflecting total navigational success combined with infinite speed and lightening computation.
    \item 0, on the other hand, corresponds to a navigation failure or either of CT and TT reaching the limit.
\end{itemize}

\paragraph{CASR Insights}
CASR serves dual purposes. Beyond being a metric for considering computational time, it acts as a performance indicator during optimization of model and hyperparameter selection. An increase in CT that doesn't correspond to a significant shift in CASR suggests that the TT remains largely unaffected by the CT changes. Further discussions on this can be found in Section \ref{dis: insightCASR}.

Note, while the notion of immediate inference or fast travel may seem far-fetched at first glance, most real-time models do operate at 30+ Hz, pushing CT $\rightarrow 0$, and setting a goal for machine learning efficiency research.  The size of TT captures route efficiency and morphology choices.  In a practical setting, one might opt for a UAV over a UGV, choose a wheeled vs legged UGV, or even optimize gait to further improve TT.  Again, as many commodity UAVs approach speeds of 100k/h, even existing hardware can for many domains shrink TT $\rightarrow 0$ if routing is correct. In our comparisons, UAV is compared to UAV and UGV to UGV so morphology does not affect comparisons, only routing.

\subsection{Baselines}
We benchmarked our approach against two baseline implementations: The first, LLM-as-Eval~\cite{chen2023train}, utilizes the LLM as an evaluator for re-scoring expanded frontiers~\cite{1997frontier}. The second, LLM-MCTS~\cite{hao2023reasoning}, employs Monte Carlo Tree Search techniques for trajectory expansion at 10 iterations. Both baselines are reproductions of existing methods using our graph maps, devoid of depth information. This choice is informed by real-world practicalities: without advanced depth cameras, standard devices such as the Real Sense D435 exhibit subpar performance in outdoor environments.

\begin{table}[t]
\centering
\begin{tabular}{@{}l@{\hspace{15pt}}c@{\hspace{10pt}}c@{\hspace{10pt}}c@{\hspace{25pt}}c@{\hspace{10pt}}c@{\hspace{10pt}}c@{}}
\toprule
            & \multicolumn{3}{c@{\hspace{25pt}}}{Simulation (Drone)} & \multicolumn{3}{@{}c}{Real World (Quadruped)} \\
            & SR & OSR & CASR & SR & OSR & CASR \\ 
\midrule
LLM-as-Eval & 0.06 & 0.06 & 0.04 & 0.10 & 0.20 & 0.04 \\
Ours       & \textbf{0.51} & \textbf{0.82} & \textbf{0.42} & \textbf{0.60} & \textbf{0.70} & \textbf{0.24} \\                             
\bottomrule
\end{tabular}
\caption{Simulation and Real World Performance.\\ CASR decreased in Real World due to use of a quadruped.}
\vspace{-20pt}
\label{tab:simvsreal}
\end{table}

\subsection{Result}

\label{sec:result}

\begin{figure}[h]
    \centering
    \includegraphics[width=0.92\linewidth]{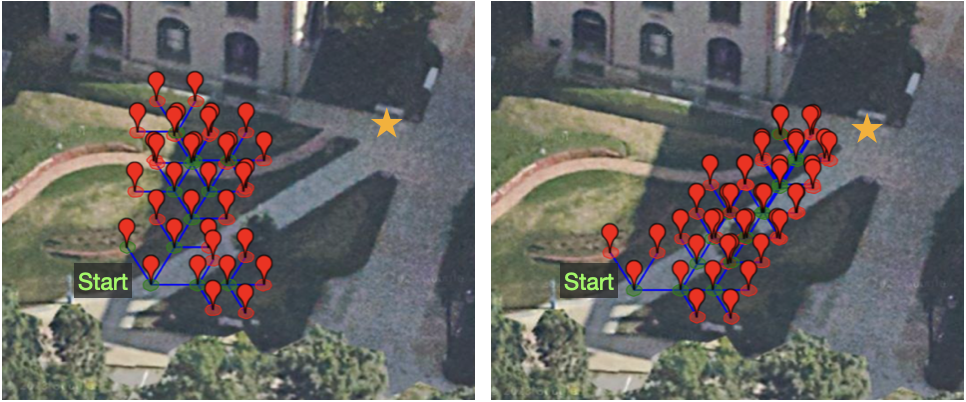}
    \caption{Comparative Trajectories of LLM-as-Eval (left) and Reasoned Explorer (right). Green nodes represent the chosen path, while red nodes highlight the frontiers.}
    \label{fig:result_vis}
\end{figure}

Table~\ref{tab:success} presents a comprehensive evaluation of our methods against the baselines on four metrics: Success Rate (SR), Oracle Success Rate (OSR), Success Weighted by Path Length (SPL)~\cite{anderson2018objectnav}, and our newly introduced metric, Computationally Adjusted Success Rate (CASR). It shows: 

\begin{itemize}
\item We consistently outperformed naive use of LLM as evaluator and LLM-MCTS across all four metrics.
\item In addition to the performance metrics, our methodology exhibited superior time efficiency compared to LLM-MCTS shown in the CASR difference. This emphasizes the practicality and efficiency of our proposed solutions in real-world applications.
\item Performance for all methods decreased as difficulty increased from L1 to L4 of our \textbf{\acro} tasks, with the exception of L3. We suspect the anomalous L3 performance is attributed to human-specific path preferences that potentially reduce the search space.
\end{itemize}

Table~\ref{tab:simvsreal} showcases the efficacy of our methods in transferring from simulation to the real-world. For comparative analysis, the baseline LLM-as-Eval was also tested in real-world scenarios to assess its potential for achieving superior performance. The results indicate that our methods smoothly transition from simulation to real-world contexts with matching performance.
Figure \ref{fig:result_vis} shows that LLM-as-Eval is more like a random search style of exploration. In contrast, our method delivers a more structured exploration towards the goal.

\subsection{Ablation Study}
\label{sec:ablation}

In our ablation study, we aimed to assess the impact of various design choices of our Reasoned Explorer method on performance. To maintain consistency, we standardized the conditions by focusing solely on a level 1 OUTDOOR task with \(T_{max}\) set to 15 minutes. Furthermore, to minimize the performance variance attributed to our perception model, we ignored the caption error in this evaluation.

\subsubsection{How many steps should we think into the future?}

\begin{wraptable}[8]{r}{0.35\linewidth}
\vspace{-10pt}
\centering
\small
\begin{tabular}{@{}lc@{\hspace{5pt}}c@{}}
      & CASR & $\sigma$\\
\toprule
L = 0 & 0.141 & 0.38\\
L = 1 & 0.516 & 0.34\\
L = 2 & \textbf{0.732} & \textbf{0.05}\\
L = 3 & 0.644 & 0.07\\
L = 4 & 0.520 & 0.34\\
\bottomrule
\end{tabular}
\caption{Choice of 
         \( L \)}
\vspace{-10pt}
\label{tab:ablate_L}
\end{wraptable}

Table~\ref{tab:ablate_L} shows how different \( L \) values affect CASR performance. At \( L\!=\!0 \), the approach aligns with the LLM-as-Eval baseline, and we subsequently increment \( L \) to 4. The optimal performance is observed at \( L\!=\!2 \), underscoring the advantages of iterative querying with LLM visionary and LLM evaluator for OUTDOOR tasks. 

Notably, the standard deviation is minimized around \( L = 2 \) and increases at the extremes. The variance at \( L = 0 \) can be attributed to relying solely on LLM's single output, while the increased uncertainty at \( L = 4 \) suggests that excessive querying might introduce performance variability. 

\begin{wraptable}[7]{r}{0.35\linewidth}
\vspace{-10pt}
\centering
\small
\begin{tabular}{@{}l@{}l@{\hspace{5pt}}c@{}}
Vis & + Eval    & CASR \\
\toprule
GPT3.5 & + 3.5 &  0.449\\
GPT3.5 & + 4 &  \textbf{0.732}\\
GPT4 & + 4 &   - \\
\bottomrule
\end{tabular}
\caption{Choice of Model}
\vspace{-10pt}
\label{tab:ablate_Model}
\end{wraptable}
\subsubsection{What LLM model should we chose?}
We conducted an ablation study focusing on the model selection for LLM\_Visionary and LLM\_Evaluator, detailed in Table \ref{tab:ablate_Model}. The pairing of GPT3.5 for visionary predictions and GPT4 for node scoring emerges as an optimal choice, balancing both efficiency and performance. The combination of GPT4 with GPT4 was not explored due to the rate limit constraints of the OpenAI API, and can be explored in future work. Notably, even with GPT3.5 serving both LLM\_Eval and LLM\_Visionary roles, wall avoidance behavior is observed. This observation could provide valuable insights for future research aiming to utilize LLM as a local planner. Another observation is that the average scores GPT3.5 as evaluator gives is 1.86 points higher than GPT4, means that GPT3.5 is more optimistic and GPT4 is more conservative in scoring.



\section{Discussions}
\label{sec:discussions}

\subsection{Obstacle Avoidance Capability}
Based on our experimental observations, our method, Reasoned Explorer, excels at navigating around larger obstacles, such as walls. However, its limitation emerges when confronted with smaller objects. The ability to avoid these objects hinges on the graph's edge length and the precision of the perception model. We posit that with a more advanced perception model, which can accurately determine the relative positions of objects, the method holds potential to adeptly handle smaller obstacles as well.

\begin{figure}[t]
    \centering
    \includegraphics[width=0.85\linewidth]{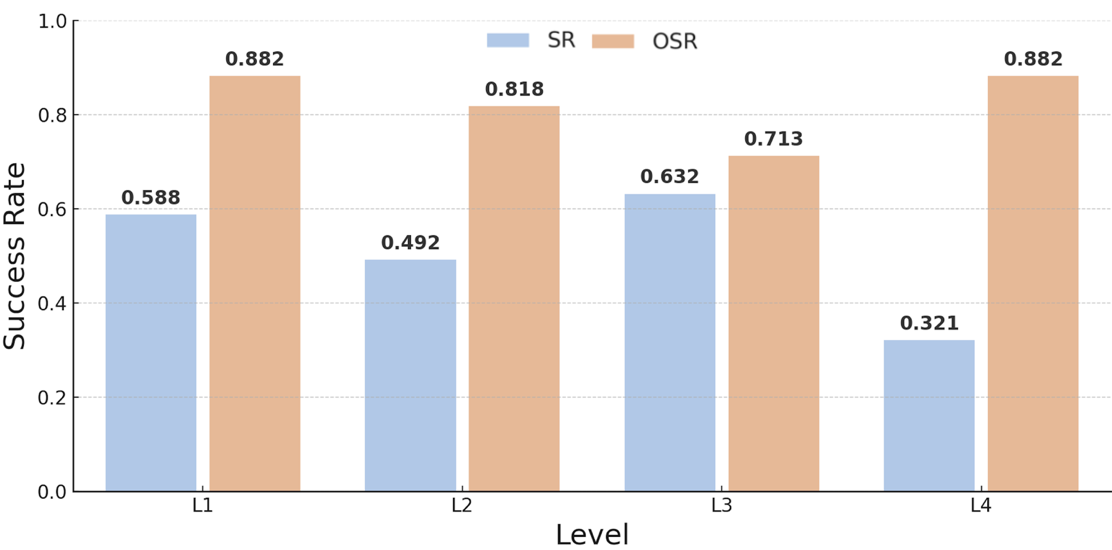}
    \caption{Comparison between SR and OSR for Reasoned Explorer}
    \label{fig:VLM_SR_OSR}
\end{figure}

\subsection{Current Limitations of VLM}
\label{dis: insightCASR}
As illustrated in Figure \ref{fig:VLM_SR_OSR}, there is a significant disparity between OSR and SR, especially with harder tasks. This divergence predominantly emerges because the VLM, upon achieving its goal, often fails to acknowledge it. This misrecognition subsequently leads the agent off target. Furthermore, the VLM has a propensity to hallucinate the positions of objects. In outdoor environments, it becomes particularly challenging to provide accurate descriptions of every object. Consequently, future work may consider the direct use of embeddings in lieu of caption data. Such observations point towards the potential of improving VLM's performance, ideally bridging the gap and enabling SR to align more closely with OSR.

\subsection{Observation Derived from CASR}

The CASR metric serves as both a metric for method comparison and a tool for understanding the balance between computational time and task efficiency. Key observations are:

\begin{enumerate}
    \item \textbf{Positive Correlation with CT}: An increase in CT leading to a higher CASR suggests that more computation can reduce travel time (TT), enhancing efficiency.
    
    \item \textbf{Negative Correlation with CT}: A drop in CASR with increased CT indicates diminishing returns from additional computation (e.g. 
    saturation).
    
    \item \textbf{Steady CASR despite CT Variation}: A consistent CASR, despite varying CT, indicates a balance between computation and task time, so 
    other factors, such as motion planning dominate. 
\end{enumerate}


\section{Conclusions}
\label{sec:conclusions}
The emergence of LLMs has opened new avenues in the embodied agent domain. This paper introduced the \textbf{\acro} task, a pioneering approach aimed at propelling embodied agents into challenging outdoor settings. 
Further, we introduced a novel, general, mechanism for using LLMs to reason about robot plans in unseen environments, 
and 
proposed the \textbf{CASR}, the first metric to assess balance between reasoning and action for embodied agents. 
Our formulation more closely mirrors how humans navigate and explore, trading off between thinking and acting to both leverage what we know in general and can see in the specific.  

\renewcommand{\bibfont}{\normalfont\small}
\printbibliography

@inproceedings{chaplot2020object,
  title={Object Goal Navigation using Goal-Oriented Semantic Exploration},
  author={Chaplot, Devendra Singh and Gandhi, Dhiraj and
            Gupta, Abhinav and Salakhutdinov, Ruslan},
  booktitle={In Neural Information Processing Systems (NeurIPS)},
  year={2020}
  }

@misc{majumdar2022zson,
      title={ZSON: Zero-Shot Object-Goal Navigation using Multimodal Goal Embeddings}, 
      author={Arjun Majumdar and Gunjan Aggarwal and Bhavika Devnani and Judy Hoffman and Dhruv Batra},
      year={2022},
      eprint={2206.12403},
      archivePrefix={arXiv},
      primaryClass={cs.CV}
}

@misc{anderson2018objectnav,
      title={On Evaluation of Embodied Navigation Agents}, 
      author={Peter Anderson and Angel Chang and Devendra Singh Chaplot and Alexey Dosovitskiy and Saurabh Gupta and Vladlen Koltun and Jana Kosecka and Jitendra Malik and Roozbeh Mottaghi and Manolis Savva and Amir R. Zamir},
      year={2018},
      eprint={1807.06757},
      archivePrefix={arXiv},
      primaryClass={cs.AI}
}

@misc{gpt,
      title={Language Models are Few-Shot Learners}, 
      author={Tom B. Brown and Benjamin Mann and Nick Ryder and Melanie Subbiah and Jared Kaplan and Prafulla Dhariwal and Arvind Neelakantan and Pranav Shyam and Girish Sastry and Amanda Askell and Sandhini Agarwal and Ariel Herbert-Voss and Gretchen Krueger and Tom Henighan and Rewon Child and Aditya Ramesh and Daniel M. Ziegler and Jeffrey Wu and Clemens Winter and Christopher Hesse and Mark Chen and Eric Sigler and Mateusz Litwin and Scott Gray and Benjamin Chess and Jack Clark and Christopher Berner and Sam McCandlish and Alec Radford and Ilya Sutskever and Dario Amodei},
      year={2020},
      eprint={2005.14165},
      archivePrefix={arXiv},
      primaryClass={cs.CL}
}

@misc{openai2023gpt4,
      title={GPT-4 Technical Report}, 
      author={OpenAI},
      year={2023},
      eprint={2303.08774},
      archivePrefix={arXiv},
      primaryClass={cs.CL}
}

@misc{wei2023chainofthought,
      title={Chain-of-Thought Prompting Elicits Reasoning in Large Language Models}, 
      author={Jason Wei and Xuezhi Wang and Dale Schuurmans and Maarten Bosma and Brian Ichter and Fei Xia and Ed Chi and Quoc Le and Denny Zhou},
      year={2023},
      eprint={2201.11903},
      archivePrefix={arXiv},
      primaryClass={cs.CL}
}

@misc{yao2023tree,
      title={Tree of Thoughts: Deliberate Problem Solving with Large Language Models}, 
      author={Shunyu Yao and Dian Yu and Jeffrey Zhao and Izhak Shafran and Thomas L. Griffiths and Yuan Cao and Karthik Narasimhan},
      year={2023},
      eprint={2305.10601},
      archivePrefix={arXiv},
      primaryClass={cs.CL}
}

@misc{xie2023decomposition,
      title={Decomposition Enhances Reasoning via Self-Evaluation Guided Decoding}, 
      author={Yuxi Xie and Kenji Kawaguchi and Yiran Zhao and Xu Zhao and Min-Yen Kan and Junxian He and Qizhe Xie},
      year={2023},
      eprint={2305.00633},
      archivePrefix={arXiv},
      primaryClass={cs.CL}
}

@inproceedings{shah2023lmnav,
  title={Lm-nav: Robotic navigation with large pre-trained models of language, vision, and action},
  author={Shah, Dhruv and Osi{\'n}ski, B{\l}a{\.z}ej and Levine, Sergey and others},
  booktitle={Conference on Robot Learning},
  pages={492--504},
  year={2023},
  organization={PMLR}
}

@inproceedings{majumdar2020vlnbert,
  title={Improving vision-and-language navigation with image-text pairs from the web},
  author={Majumdar, Arjun and Shrivastava, Ayush and Lee, Stefan and Anderson, Peter and Parikh, Devi and Batra, Dhruv},
  booktitle={Computer Vision--ECCV 2020: 16th European Conference, Glasgow, UK, August 23--28, 2020, Proceedings, Part VI 16},
  pages={259--274},
  year={2020},
  organization={Springer}
}

@article{devlin2018bert,
  title={Bert: Pre-training of deep bidirectional transformers for language understanding},
  author={Devlin, Jacob and Chang, Ming-Wei and Lee, Kenton and Toutanova, Kristina},
  journal={arXiv preprint arXiv:1810.04805},
  year={2018}
}

@article{dorbala2023catshapedmug,
  title={Can an embodied agent find your" cat-shaped mug"? llm-based zero-shot object navigation},
  author={Dorbala, Vishnu Sashank and Mullen Jr, James F and Manocha, Dinesh},
  journal={arXiv preprint arXiv:2303.03480},
  year={2023}
}

@article{yao2022react,
  title={React: Synergizing reasoning and acting in language models},
  author={Yao, Shunyu and Zhao, Jeffrey and Yu, Dian and Du, Nan and Shafran, Izhak and Narasimhan, Karthik and Cao, Yuan},
  journal={arXiv preprint arXiv:2210.03629},
  year={2022}
}

@article{brohan2022rt,
  title={Rt-1: Robotics transformer for real-world control at scale},
  author={Brohan, Anthony and Brown, Noah and Carbajal, Justice and Chebotar, Yevgen and Dabis, Joseph and Finn, Chelsea and Gopalakrishnan, Keerthana and Hausman, Karol and Herzog, Alex and Hsu, Jasmine and others},
  journal={arXiv preprint arXiv:2212.06817},
  year={2022}
}

@misc{chen2023train,
    title={How To Not Train Your Dragon: Training-free Embodied Object Goal Navigation with Semantic Frontiers}, 
    author={Junting Chen and Guohao Li and Suryansh Kumar and Bernard Ghanem and Fisher Yu},
    year={2023},
    eprint={2305.16925},
    archivePrefix={arXiv},
    primaryClass={cs.CV}
}

@misc{zhou2023esc,
      title={ESC: Exploration with Soft Commonsense Constraints for Zero-shot Object Navigation}, 
      author={Kaiwen Zhou and Kaizhi Zheng and Connor Pryor and Yilin Shen and Hongxia Jin and Lise Getoor and Xin Eric Wang},
      year={2023},
      eprint={2301.13166},
      archivePrefix={arXiv},
      primaryClass={cs.AI}
}

@misc{zhou2023navgpt,
      title={NavGPT: Explicit Reasoning in Vision-and-Language Navigation with Large Language Models}, 
      author={Gengze Zhou and Yicong Hong and Qi Wu},
      year={2023},
      eprint={2305.16986},
      archivePrefix={arXiv},
      primaryClass={cs.CV}
}

@article{hao2023reasoning,
  title={Reasoning with language model is planning with world model},
  author={Hao, Shibo and Gu, Yi and Ma, Haodi and Hong, Joshua Jiahua and Wang, Zhen and Wang, Daisy Zhe and Hu, Zhiting},
  journal={arXiv preprint arXiv:2305.14992},
  year={2023}
}

@inproceedings{
zhang2023planning,
title={Planning with Large Language Models for Code Generation},
author={Shun Zhang and Zhenfang Chen and Yikang Shen and Mingyu Ding and Joshua B. Tenenbaum and Chuang Gan},
booktitle={The Eleventh International Conference on Learning Representations },
year={2023},
url={https://openreview.net/forum?id=Lr8cOOtYbfL}
}

@misc{wang2023describe,
      title={Describe, Explain, Plan and Select: Interactive Planning with Large Language Models Enables Open-World Multi-Task Agents}, 
      author={Zihao Wang and Shaofei Cai and Anji Liu and Xiaojian Ma and Yitao Liang},
      year={2023},
      eprint={2302.01560},
      archivePrefix={arXiv},
      primaryClass={cs.AI}
}

@misc{wang2023planandsolve,
      title={Plan-and-Solve Prompting: Improving Zero-Shot Chain-of-Thought Reasoning by Large Language Models}, 
      author={Lei Wang and Wanyu Xu and Yihuai Lan and Zhiqiang Hu and Yunshi Lan and Roy Ka-Wei Lee and Ee-Peng Lim},
      year={2023},
      eprint={2305.04091},
      archivePrefix={arXiv},
      primaryClass={cs.CL}
}

@misc{yao2023react,
      title={ReAct: Synergizing Reasoning and Acting in Language Models}, 
      author={Shunyu Yao and Jeffrey Zhao and Dian Yu and Nan Du and Izhak Shafran and Karthik Narasimhan and Yuan Cao},
      year={2023},
      eprint={2210.03629},
      archivePrefix={arXiv},
      primaryClass={cs.CL}
}

@inproceedings{huang2022inner,
    title={Inner Monologue: Embodied Reasoning through Planning with Language Models},
    author={Wenlong Huang and Fei Xia and Ted Xiao and Harris Chan and Jacky Liang and Pete Florence and Andy Zeng and Jonathan Tompson and Igor Mordatch and Yevgen Chebotar and Pierre Sermanet and Noah Brown and Tomas Jackson and Linda Luu and Sergey Levine and Karol Hausman and Brian Ichter},
    booktitle={arXiv preprint arXiv:2207.05608},
    year={2022}
}

@inproceedings{
zhao2023large,
title={Large Language Models as Commonsense Knowledge for Large-Scale Task Planning},
author={Zirui Zhao and Wee Sun Lee and David Hsu},
booktitle={RSS 2023 Workshop on Learning for Task and Motion Planning},
year={2023},
url={https://openreview.net/forum?id=tED747HURfX}
}

@misc{schumann2023velma,
      title={VELMA: Verbalization Embodiment of LLM Agents for Vision and Language Navigation in Street View}, 
      author={Raphael Schumann and Wanrong Zhu and Weixi Feng and Tsu-Jui Fu and Stefan Riezler and William Yang Wang},
      year={2023},
      eprint={2307.06082},
      archivePrefix={arXiv},
      primaryClass={cs.AI}
}

@INPROCEEDINGS{1997frontier,
  author={Yamauchi, B.},
  booktitle={Proceedings 1997 IEEE International Symposium on Computational Intelligence in Robotics and Automation CIRA'97. 'Towards New Computational Principles for Robotics and Automation'}, 
  title={A frontier-based approach for autonomous exploration}, 
  year={1997},
  volume={},
  number={},
  pages={146-151},
  doi={10.1109/CIRA.1997.613851}}

@article{kosmos-2,
  title={Kosmos-2: Grounding Multimodal Large Language Models to the World},
  author={Zhiliang Peng and Wenhui Wang and Li Dong and Yaru Hao and Shaohan Huang and Shuming Ma and Furu Wei},
  journal={ArXiv},
  year={2023},
  volume={abs/2306}
}

@misc{shah2017airsim,
      title={AirSim: High-Fidelity Visual and Physical Simulation for Autonomous Vehicles}, 
      author={Shital Shah and Debadeepta Dey and Chris Lovett and Ashish Kapoor},
      year={2017},
      eprint={1705.05065},
      archivePrefix={arXiv},
      primaryClass={cs.RO}
}

@inproceedings{saycan2022arxiv,
    title={Do As I Can and Not As I Say: Grounding Language in Robotic Affordances},
    author={Michael Ahn and Anthony Brohan and Noah Brown and Yevgen Chebotar and Omar Cortes and Byron David and Chelsea Finn and Chuyuan Fu and Keerthana Gopalakrishnan and Karol Hausman and Alex Herzog and Daniel Ho and Jasmine Hsu and Julian Ibarz and Brian Ichter and Alex Irpan and Eric Jang and Rosario Jauregui Ruano and Kyle Jeffrey and Sally Jesmonth and Nikhil Joshi and Ryan Julian and Dmitry Kalashnikov and Yuheng Kuang and Kuang-Huei Lee and Sergey Levine and Yao Lu and Linda Luu and Carolina Parada and Peter Pastor and Jornell Quiambao and Kanishka Rao and Jarek Rettinghouse and Diego Reyes and Pierre Sermanet and Nicolas Sievers and Clayton Tan and Alexander Toshev and Vincent Vanhoucke and Fei Xia and Ted Xiao and Peng Xu and Sichun Xu and Mengyuan Yan and Andy Zeng},
    booktitle={arXiv preprint arXiv:2204.01691},
    year={2022}
}

@inproceedings{blind-indoors,
author = {Nanavati, Amal and Tan, Xiang Zhi and Steinfeld, Aaron},
title = {Coupled Indoor Navigation for People Who Are Blind},
year = {2018},
isbn = {9781450356152},
publisher = {Association for Computing Machinery},
address = {New York, NY, USA},
url = {https://doi.org/10.1145/3173386.3176976},
doi = {10.1145/3173386.3176976},
booktitle = {Companion of the 2018 ACM/IEEE International Conference on Human-Robot Interaction},
pages = {201–202},
numpages = {2},
location = {Chicago, IL, USA},
series = {HRI '18}
}

@inproceedings{chaplot2020learning,
  title={Learning To Explore Using Active Neural SLAM},
  author={Chaplot, Devendra Singh and Gandhi, Dhiraj and Gupta,
          Saurabh and Gupta, Abhinav and Salakhutdinov, Ruslan},
  booktitle={International Conference on 
             Learning Representations (ICLR)},
  year={2020}}

@inproceedings{Min2022,
    author    = {So Yeon Min and Devendra Singh Chaplot and Pradeep Ravikumar and Yonatan Bisk and Ruslan Salakhutdinov},
    title     = {{FILM: Following Instructions in Language with Modular Methods}},
    booktitle = {The Tenth International Conference on Learning Representations},
    year      = {2022},
    url       = {https://soyeonm.github.io/FILM_webpage/},
}

@INPROCEEDINGS{self-supervised-outdoor,

  author={Castro, Mateo Guaman and Triest, Samuel and Wang, Wenshan and Gregory, Jason M. and Sanchez, Felix and Rogers, John G. and Scherer, Sebastian},

  booktitle={2023 IEEE International Conference on Robotics and Automation (ICRA)}, 

  title={How Does It Feel? Self-Supervised Costmap Learning for Off-Road Vehicle Traversability}, 

  year={2023},

  volume={},

  number={},

  pages={931-938},

  doi={10.1109/ICRA48891.2023.10160856}}

@InProceedings{pmlr-v155-anderson21a,
  title = 	 {Sim-to-Real Transfer for Vision-and-Language Navigation},
  author =       {Anderson, Peter and Shrivastava, Ayush and Truong, Joanne and Majumdar, Arjun and Parikh, Devi and Batra, Dhruv and Lee, Stefan},
  booktitle = 	 {Proceedings of the 2020 Conference on Robot Learning},
  pages = 	 {671--681},
  year = 	 {2021},
  editor = 	 {Kober, Jens and Ramos, Fabio and Tomlin, Claire},
  volume = 	 {155},
  series = 	 {Proceedings of Machine Learning Research},
  month = 	 {16--18 Nov},
  publisher =    {PMLR},
  url = 	 {https://proceedings.mlr.press/v155/anderson21a.html}
}

@article{McKenna2023SourcesOH,
  title={Sources of Hallucination by Large Language Models on Inference Tasks},
  author={Nick McKenna and Tianyi Li and Liang Cheng and Mohammad Javad Hosseini and Mark Johnson and Mark Steedman},
  journal={ArXiv},
  year={2023},
  volume={abs/2305.14552},
}

@inproceedings{stochastic,
author = {Bender, Emily M. and Gebru, Timnit and McMillan-Major, Angelina and Shmitchell, Shmargaret},
title = {On the Dangers of Stochastic Parrots: Can Language Models Be Too Big?},
year = {2021},
isbn = {9781450383097},
publisher = {Association for Computing Machinery},
address = {New York, NY, USA},
booktitle = {Proceedings of the 2021 ACM Conference on Fairness, Accountability, and Transparency},
pages = {610–623},
numpages = {14},
location = {Virtual Event, Canada},
series = {FAccT '21}
}
\end{document}